\documentclass[sigconf, anonymous=false]{acmart}
\AtBeginDocument{%
  \providecommand\BibTeX{{%
    Bib\TeX}}}

\copyrightyear{2024}
\acmYear{2024}
\setcopyright{acmlicensed}\acmConference[AutonomousCyber '24]{Proceedings of the Workshop on Autonomous Cybersecurity}{October 14--18, 2024}{Salt Lake City, UT, USA}
\acmBooktitle{Proceedings of the Workshop on Autonomous Cybersecurity (AutonomousCyber '24), October 14--18, 2024, Salt Lake City, UT, USA}
\acmDOI{10.1145/3689933.3690833}
\acmISBN{979-8-4007-1229-6/24/10}
\usepackage[utf8]{inputenc}
\usepackage{bm} 

\usepackage{bbding} 
\usepackage{multirow}

\usepackage[ruled,linesnumbered]{algorithm2e}

\SetAlCapNameFnt{\small}
\SetAlCapFnt{\small}

\SetCommentSty{mycommfont}

\microtypecontext{spacing=nonfrench}

\usepackage{adjustbox}
\usepackage{booktabs}
\usepackage{circledsteps}
\usepackage{paralist}
\usepackage[group-digits=integer,group-minimum-digits=4,group-separator={{,}},per-mode=symbol,detect-all]{siunitx}
\usepackage{xspace}
\usepackage{tcolorbox}

\hypersetup{
     colorlinks=true,
     citecolor=blue,
     linkcolor=black,
}

\newif\ifDraft

\Draftfalse

\usepackage[lambda,adversary,asymptotics,sets,landau,probability,operators]{cryptocode}

\makeatletter
\g@addto@macro{\UrlBreaks}{\UrlOrds}
\makeatother

\usepackage[lambda,adversary,asymptotics,sets,landau,probability,operators]{cryptocode}
\usepackage{cleveref}

\bgroup\fboxsep=0pt

\NewEnviron{dottedbox}{
\par
\begin{tikzpicture}
\node[rectangle,minimum width=0.9\textwidth] (m) {\begin{minipage}{\textwidth}\BODY\end{minipage}};
\draw[dashed] (m.south west) rectangle (m.north east);
\end{tikzpicture}
}

\definecolor{gamechangecolor}{gray}{0.90}

\tcbuselibrary{theorems}
\tcbuselibrary{skins}
\makeatletter
\crefformat{tcb@cnt@protocol}{protocol~#2#1#3}
\Crefformat{tcb@cnt@Protocol}{Protocol~#2#1#3}

\makeatother
\newtcbtheorem[auto counter]{protocol}{Protocol}{
	left=8pt,
	right=8pt,
	top=0pt,
	bottom=0pt,
    fonttitle=\normalsize,
    fontupper=\footnotesize,
	coltitle=black!,
	colbacktitle=black!0,
	colback=black!0, 
	boxrule=0.5pt,
	colframe=black!100,
	coltext=black!100,
	arc=0pt
}{proto}

\makeatother
\newtcbtheorem[auto counter]{functionality}{Functionality}{
	left=8pt,
	right=8pt,
	top=0pt,
	bottom=0pt,
    fonttitle=\bfseries\normalsize,
    fontupper=\small,
	coltitle=black!,
	colbacktitle=black!0,
	colback=black!0, 
	boxrule=0.5pt,
	colframe=black!100,
	coltext=black!100,
	arc=0pt
}{func}

\crefformat{tcb@cnt@functionality}{functionality~#2#1#3}
\Crefformat{tcb@cnt@Functionality}{Functionality~#2#1#3}

\usepackage{seqsplit}
\newcommand{\ttt}[1]{%
  \begingroup
    \protect\renewcommand{\seqinsert}{\ifmmode\allowbreak\else\-\fi}%
    \protect\texttt{\protect\seqinsert{\protect\seqsplit{\small#1}}}%
  \endgroup
}
\newcommand{\tttscript}[1]{%
  \begingroup
    \protect\renewcommand{\seqinsert}{\ifmmode\allowbreak\else\-\fi}%
    \protect\texttt{\protect\seqinsert{\protect\seqsplit{\scriptsize#1}}}%
  \endgroup
}
\newcommand{\tttfoot}[1]{%
  \begingroup
    \protect\renewcommand{\seqinsert}{\ifmmode\allowbreak\else\-\fi}%
    \protect\texttt{\protect\seqinsert{\protect\seqsplit{\footnotesize#1}}}%
  \endgroup
}

\usetikzlibrary{shapes, automata, positioning, arrows, calc}
\tikzset{
    >=stealth, 
    node distance=3.5cm, 
    every state/.style={rectangle, thick, fill=gray!10}, 
    initial text=$ $, 
}

\colorlet{myred}{red!50!white}

\def\BibTeX{{\rm B\kern-.05em{\sc i\kern-.025em b}\kern-.08em T\kern-.1667em\lower.7ex\hbox{E}\kern-.125emX}}
\title[Towards Autonomous Cybersecurity: ...]{Towards Autonomous Cybersecurity: An Intelligent AutoML Framework for Autonomous Intrusion Detection}
\author{Li Yang}
\orcid{0000-0001-9383-1097}
 \affiliation{%
   \institution{Ontario Tech University}
   \city{Oshawa}
   \country{Canada} }
\email{li.yang@ontariotechu.ca}

\author{Abdallah Shami}
 \affiliation{%
   \institution{Western University}
   \city{London}
   \country{Canada} }
\email{abdallah.shami@uwo.ca}

\begin{document}

\begin{abstract}
The rapid evolution of mobile networks from 5G to 6G has necessitated the development of autonomous network management systems, such as Zero-Touch Networks (ZTNs). However, the increased complexity and automation of these networks have also escalated cybersecurity risks. Existing Intrusion Detection Systems (IDSs) leveraging traditional Machine Learning (ML) techniques have shown effectiveness in mitigating these risks, but they often require extensive manual effort and expert knowledge. To address these challenges, this paper proposes an Automated Machine Learning (AutoML)-based autonomous IDS framework towards achieving autonomous cybersecurity for next-generation networks. To achieve autonomous intrusion detection, the proposed AutoML framework automates all critical procedures of the data analytics pipeline, including data pre-processing, feature engineering, model selection, hyperparameter tuning, and model ensemble. Specifically, it utilizes a Tabular Variational Auto-Encoder (TVAE) method for automated data balancing, tree-based ML models for automated feature selection and base model learning, Bayesian Optimization (BO) for hyperparameter optimization, and a novel Optimized Confidence-based Stacking Ensemble (OCSE) method for automated model ensemble. The proposed AutoML-based IDS was evaluated on two public benchmark network security datasets, CICIDS2017 and 5G-NIDD, and demonstrated improved performance compared to state-of-the-art cybersecurity methods. This research marks a significant step towards fully autonomous cybersecurity in next-generation networks, potentially revolutionizing network security applications.
\end{abstract}


\maketitle

\section{Introduction}
The progression of mobile networks has played a pivotal role in the digital revolution, with each generation bringing forth new technologies and capabilities. The fifth-generation (5G) networks have significantly enhanced mobile broadband and enabled massive machine-type communications with ultra-reliable low latency \cite{6g1}. 5G networks leverages abstraction and virtualization techniques, such as Software-Defined Networking (SDN), Network Function Virtualization (NFV), and Network Slicing (NS), to provide flexible, efficient, and automated network management and services \cite{5g}.

For the evolution from 5G to the sixth generation (6G) networks, network automation has become a necessity to meet the unprecedented demand for future network applications. 6G networks are expected to leverage Artificial Intelligence (AI), Machine Learning (ML), and automation techniques to provide functional modules and operational services, leading to self-organizing and autonomous networks \cite{6g1}. Previous researchers have extensively dedicated efforts to developing network automation architectures, including Intent-Based Network Management (IBN), Self-Organizing Network Management (SON), and Autonomic Network Management (ANM), etc. \cite{zsm0}. Recently, Zero-Touch Networks (ZTNs) were proposed by the European Telecommunications Standards Institute (ETSI) as a fully autonomous network management architecture with minimal human involvement \cite{zsm}. Network automation solutions, including ZTNs, can effectively decrease network operational costs, enhance resource utilization efficiency, and mitigate the risks associated with human errors. 

On the other hand, as network and service management requires a trustworthy and reliable system, cybersecurity has become a critical component of next-generation networks. Modern networks are vulnerable to various cyber-attacks, such as Denial of Service (DoS), sniffing/eavesdropping, spoofing, web attacks, and botnets \cite{mth}. These threats can lead to severe consequences, including financial losses, disruption of critical services, compromise of sensitive information, and reputational damage \cite{sec1}. Therefore, effective cybersecurity measures should be developed to enhance the security of modern networks, while autonomous cybersecurity solutions are essential for safeguarding future networks with high automation requirements.

AI/ML techniques are widely used in network applications to develop data-driven cybersecurity mechanisms such as Intrusion Detection Systems (IDSs) and anomaly detection systems, which can analyze network traffic patterns and identify anomalies or cyber-attacks \cite{idsml1}. AI/ML models have shown effectiveness in network data analytics and IDS development, due to their capability to large volumes of network data, identify complex patterns, and adapt to evolving threats. ML-based IDSs can detect malicious attacks and predict potential threats based on historical data, thereby triggering countermeasures or response mechanisms to safeguard against the detected attacks \cite{mth}. 

To ensure robust cybersecurity in next-generation networks, such as ZTNs, it is crucial to incorporate self-management functionalities that address security concerns, such as self-configuration, self-monitoring, self-healing, self-protection, and self-optimization \cite{zsm2}. To meet these requirements, autonomous cybersecurity solutions, such as autonomous IDSs, should be developed to automatically monitor network activities, detect network anomalies, and identify potential attacks.

Automated ML (AutoML) techniques, which are developed to automate the design and implementation of ML models, are promising solutions to realize network automation for ZTNs or future networks \cite{myautoml}. AutoML techniques offer the advantage of automating laborious and repetitive tasks involved in the ML and data analytics pipeline, such as data pre-processing, feature engineering, model selection, and hyperparameter tuning \cite{myautoml}. This automation can effectively reduce human effort, minimize the occurrence of human errors, and alleviate the need for extensive expert knowledge. In the cybersecurity domain, autonomous IDSs can be developed using AutoML techniques by automatically designing, tuning, and optimizing ML models that can effectively detect cyber-attacks and achieve self-monitoring and self-protection.

Therefore, this paper proposes an AutoML-based autonomous IDS framework to automatically detect malicious cyber-attacks for safeguading 5G and potential 6G networks. The proposed AutoML framework enables the automation of critical procedures of the ML/data analytics pipeline for intrusion detection. Specifically, it consists of: an Automated Data Pre-processing (AutoDP) component that focuses on automated data balancing using the Tabular Variational Auto-Encoder (TVAE) \cite{tvae} method to address class-imbalance issues and improve data quality, an Automated Feature Engineering (AutoFE) component that automatically selects the most relevant features based on their average importance scores calculated using the Gini index and entropy metrics, an automated base model learning and selection component that automatically trains six tree-based machine learning models—Decision Tree (DT) \cite{mth}, Random Forest (RF) \cite{rf}, Extra Trees (ET) \cite{et}, Extreme Gradient Boosting (XGBoost) \cite{xgb}, Light Gradient Boosting Machine (LightGBM) \cite{lgb}, and Categorical Boosting (CatBoost) \cite{cb}—and selects the top three best-performing models from them, a Hyper-Parameter Optimization (HPO) component that automatically tunes and optimizes the hyperparameters of the selected ML models using Bayesian Optimization with Tree-structured Parzen Estimator (BO-TPE) \cite{bo2} to obtain optimized base models, and an automated model ensemble component that employs the proposed novel Optimized Confidence-based Stacking Ensemble (OCSE) method to generate the meta-learner for final intrusion detection. Overall, the proposed AutoML-based IDS can automatically process network data and generate optimized ML models capable of detecting various types of cyber-attacks to safeguard current and future networks.

This paper presents the following key contributions:
\begin{enumerate}
\item It proposes a novel and comprehensive AutoML framework\footnote{
Code for this paper is publicly available at: \url{https://github.com/Western-OC2-Lab/AutonomousCyber-AutoML-based-Autonomous-Intrusion-Detection-System}}. that enables fully autonomous intrusion detection in next-generation networks, holding the potential to achieve fully autonomous cybersecurity. 
\item It proposes a novel automated data balancing method based on TVAE and class distribution exploration. 
\item It proposes a novel ensemble learning method, OCSE, which extends the traditional stacking ensemble method by incorporating confidence values of classes and the BO-TPE method for model optimization. 
\item It assesses the proposed AutoML-based IDS model using two public benchmark network security datasets, CICIDS2017 \cite{cic} and 5G-NIDD \cite{5gnidd} datasets, which contain state-of-the-art cyber-attack scenarios.
\item It compares the performance of the proposed AutoML-based IDS model with state-of-the-art methods. 
\end{enumerate}


To the best of our knowledge, no previous research has proposed such a comprehensive autonomous IDS model that leverages AutoML to automate all essential network data analytics procedures, ensuring efficient and automatic detection of diverse cyber-attacks for safeguarding 5G and next-generation networks.

The paper is structured as follows: Section 2 introduces the related work using AI/ML and AutoML-based methods for developing IDSs and cybersecurity mechanisms. Section 3 presents a detailed description of the proposed AutoML-based IDS framework, including AutoDP, AutoFE, automated base model selection, HPO, and automated model ensemble. Section 4 presents and discusses the experimental results of evaluating the proposed framework on benchmark network datasets. Finally, Section 5 summarizes the paper.

\section{Related Work}
\label{S2}
AI/ML models have been extensively applied in recent years to the development of IDSs for modern networks. This related work section aims to provide an overview of the critical studies that have contributed to the development and advancement of IDSs using ML and AutoML models for future networks.

\begin{table*}[tbp]
\caption{Comparison of Various IDS Approaches with Emphasis on ML and AutoML Components.}
\label{table:comparison}
\centering
\setlength\extrarowheight{1pt}
\scalebox{0.84}{
\begin{tabular}{|>{\centering\arraybackslash}m{9.3em}|>{\centering\arraybackslash}m{7em}|>{\centering\arraybackslash}m{7em}|>{\centering\arraybackslash}m{5.5em}|>{\centering\arraybackslash}m{4.7em}|>{\centering\arraybackslash}m{4.7em}|>{\centering\arraybackslash}m{6.8em}|>{\centering\arraybackslash}m{5.5em}|>{\centering\arraybackslash}m{5.5em}|}
\hline
\textbf{Paper} & \textbf{Benchmark Dataset Evaluation} & \textbf{Traditional ML Models} & \textbf{DL Models} & \textbf{AutoDP} & \textbf{AutoFE} & \textbf{Model Optimization} & \textbf{Automated Model Selection} & \textbf{Model Ensemble} \\ \hline
Sharafaldin \textit{et al.} \cite{cic} & \Checkmark & \Checkmark & \Checkmark & & & & & \\ \hline
Maseer \textit{et al.} \cite{ml1} & \Checkmark & \Checkmark & \Checkmark & & & & & \\ \hline
Yang \textit{et al.} \cite{mth} & \Checkmark & \Checkmark & & & & \Checkmark & & \Checkmark \\ \hline
Agrafiotis \textit{et al.} \cite{ids2} & \Checkmark & & \Checkmark & & & & & \\ \hline
Tayfour \textit{et al.} \cite{ids3} & \Checkmark & & \Checkmark & & & & & \\ \hline
He \textit{et al.} \cite{ids4} & \Checkmark & & \Checkmark & \Checkmark & & \Checkmark & & \\ \hline
Khan \textit{et al.} \cite{ids5} & \Checkmark & \Checkmark & \Checkmark & & \Checkmark & \Checkmark & \Checkmark & \Checkmark \\ \hline
Elmasry \textit{et al.} \cite{ids6} & \Checkmark & & \Checkmark & & \Checkmark & \Checkmark & \Checkmark & \\ \hline
Singh \textit{et al.} \cite{ids7} & \Checkmark & \Checkmark & & & & \Checkmark & \Checkmark & \Checkmark \\ \hline
Proposed AutoML Framework & \Checkmark & \Checkmark & \Checkmark & \Checkmark & \Checkmark & \Checkmark & \Checkmark & \Checkmark \\ \hline
\end{tabular}
}
\end{table*}

\subsection{AI/ML-based IDSs}
Research on developing IDSs using AI/ML models has gained significant attention and importance, as threat hunting and cyber-attack detection are critical components of cybersecurity systems for modern networks.

Traditional AI/ML algorithms have demonstrated their effectiveness in intrusion detection, especially tree-based algorithms such as DT and RF. Sharafaldin \textit{et al.} \cite{cic} created the benchmark network security dataset, CICIDS2017, and observed that the DT and RF algorithms outperformed the other compared ML models on this dataset. Maseer \textit{et al.} \cite{ml1} proposed a ML-based benchmarking Anomaly-based IDS (AIDS) approach that develops ten typical supervised and unsupervised ML models and evaluates their performance on the CICIDS2017 dataset. The experimental results illustrate that the DT and K-Nearest Neighbor (KNN) based AIDS models perform the best on the CICIDS2017 dataset among the evaluated ML models. Yang \textit{et al.} \cite{mth} proposed a Multi-Tiered Hybrid IDS (MTH-IDS) framework for intrusion detection in vehicular networks. It incorporates both supervised learning algorithms (DT, RF, ET, and XGBoost) and unsupervised learning methods (k-means) to detect multiple types of cyber-attacks. They evaluated their framework on the CAN-intrusion dataset and the CICIDS2017 dataset to emphasize the model's effectiveness.

The utilization of Deep Learning (DL) methods in the development of IDSs has become prevalent due to their effectiveness in handling high-dimensional network traffic data. Agrafiotis \textit{et al.} \cite{ids2} proposed the embeddings and Fully-Connected network (Embeddings \& FC) model to detect malware traffic in 5G networks. This IDS model employs the Long Short-Term Memory Autoencoders (LSTM-AE) to transform packets into embeddings and uses the Fully-Connected (FC) network model to identify attacks. The Embeddings \& FC IDS demonstrates improved accuracy when applied to the 5G-NIDD dataset, a dedicated dataset for 5G networks. Tayfour \textit{et al.} \cite{ids3} proposed a DL-LSTM method supported by Software-Defined Networking (SDN) to detect cyber-attacks in the Internet of Things (IoT) and 5G networks. The DL-LSTM model achieved high accuracy on the CICIDS2017 dataset, demonstrating the effectiveness of deep learning in network intrusion detection. He \textit{et al.} \cite{ids4} proposed a Pyramid Depthwise Separable Convolution neural network-based IDS (PyDSC-IDS) for network intrusion detection. The PyDSC-IDS model uses Pyramid convolution (PyConv) to extract features from data and Depthwise Separable Convolution (DSC) to reduce model complexity. Compared with other DL models, PyDSC-IDS achieves higher detection accuracy with only a small increase in complexity on the NSL-KDD, UNSW-NB15, and CICDIDS2017 datasets.

Due to the robustness of tree-based ML algorithms in handling large-scale, high-dimensional, and non-linear network data, they are utilized as base models in the proposed framework for intrusion detection. While DL models offer powerful data analysis capabilities, they often come with higher computational complexity compared to traditional ML algorithms. To mitigate the impact of this challenge, the proposed framework employs the TVAE method, a DL model, only for synthesizing samples for minority classes. This approach proves to be more efficient than using it for intrusion detection, as processing minority class samples is significantly faster than dealing with the entire large dataset. Furthermore, the development of traditional ML/DL models for intrusion detection poses several critical challenges, such as manual effort, human bias \& errors, and expertise requirements. These challenges underscore the importance of automating AI/ML models and developing autonomous IDSs.

\subsection{AutoML-based IDSs}
AutoML techniques are promising solutions to develop autonomous IDSs by automating the tedious procedures in the data analytics/ML pipeline. While AutoML is a relatively new research area in IDS development, several recent works have already employed AutoML techniques to create autonomous IDSs for modern networks. Yang \textit{et al.} \cite{myautoml} provided a comprehensive discussion on the general and specific procedures of applying AutoML techniques to IoT data analytics and conducted a case study to employ AutoML for IoT intrusion detection tasks. Khan \textit{et al.} \cite{ids5} proposed an Optimized Ensemble IDS (OE-IDS) for intrusion detection in network environments. It automates the hyperparameter tuning process of four supervised ML algorithms and uses them to develop an ensemble model based on a soft-voting method. The OE-IDS model achieved better accuracy and F1-scores than most other compared traditional ML models on the CICIDS2017 and UNSW-NB15 datasets. Elmasry \textit{et al.} \cite{ids6} proposed a double PSO and DL-based IDS for network intrusion detection. It involves the Particle Swarm Optimization (PSO) method to select features and tune hyperparameters of three DL methods: Deep Neural Networks (DNN), LSTM, and Deep Belief Networks (DBN). This IDS model outperforms other compared methods in terms of accuracy and detection rate on the CICIDS2017 dataset. Singh \textit{et al.} \cite{ids7} proposed AutoML-ID, an AutoML-based IDS designed for Wireless Sensor Networks (WSNs). The AutoML-ID approach focuses on simple automated ML model selection and hyperparameter optimization using Bayesian Optimization (BO). The model AutoML-ID was tested on a public IDS dataset, Intrusion-Data-WSN, and achieved better performance than traditional ML models. 

The existing literature has demonstrated the advantages of AutoML-based IDSs in improving performance and reducing human effort in intrusion detection and cybersecurity applications. However, many current AutoML-based IDS models only focus on automated model selection and hyperparameter optimization, leaving significant potential for improvement in other crucial stages of the AutoML pipeline. In our proposed AutoML framework, we aim to propose and develop techniques to automate every critical step in the data analytics pipeline, including the TVAE-based automated data balancing method in the AutoDP process to handle class imbalance issues, the tree-based averaging method in the AutoFE process to reduce noise and data complexity, automated model selection and BO-based hyperparameter optimization on tree-based ML algorithms to acquire optimized base models, and the proposed OCSE method for automated model ensemble to further enhance model performance. Table \ref{table:comparison} summarizes and compares the contributions of existing literature introduced in Section \ref{S2}. 

Overall, this paper presents a generic, comprehensive, and fully automated AutoML framework for future networks with high automation requirements.

\section{Proposed Framework}
\label{S3}

\begin{figure}
     \centering
     \includegraphics[width=8.1cm]{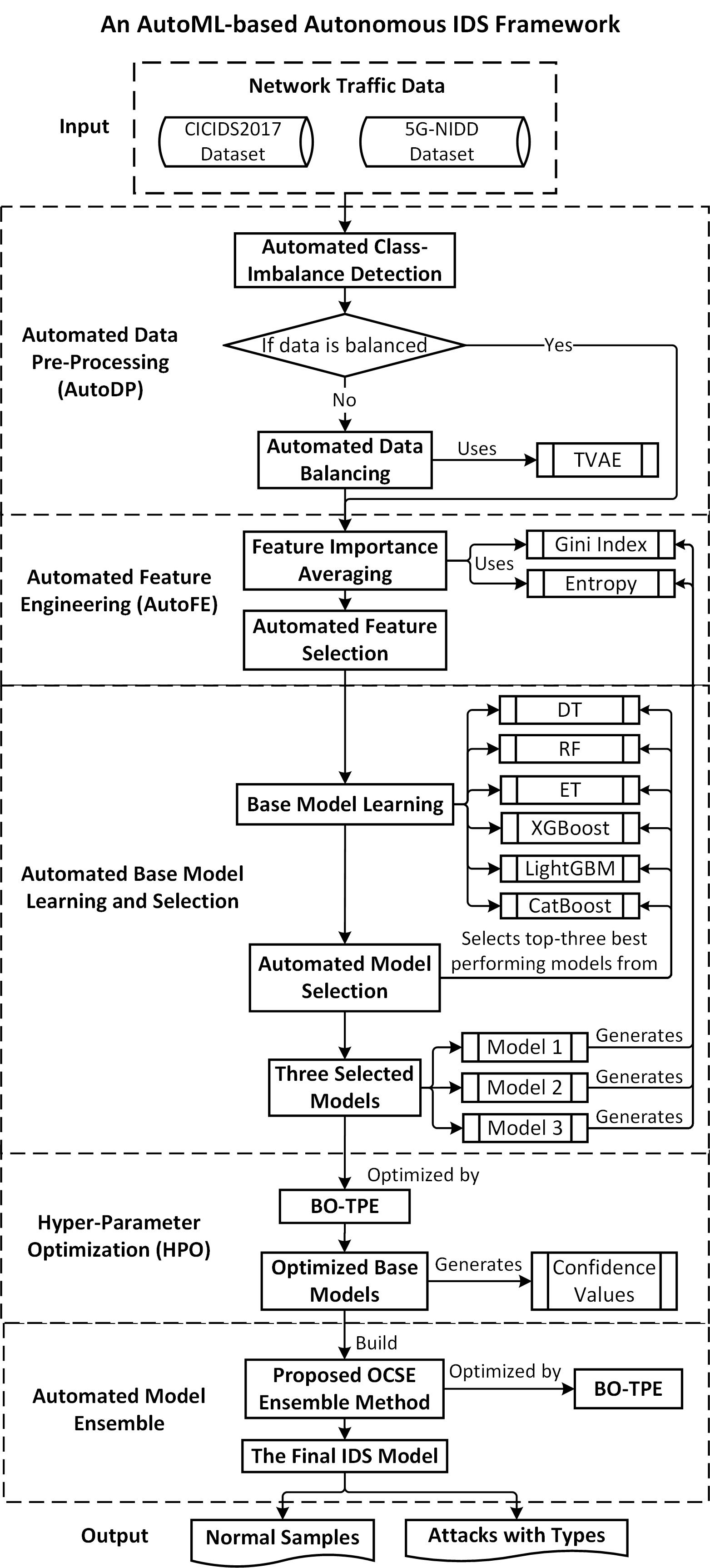}
     \caption{The proposed AutoML-based IDS framework.} 
     \label{framework}
\end{figure}

\subsection{System Overview}
The objective of this study is to develop an autonomous IDS model capable of detecting various cyber-attacks to safeguard 5G and potential 6G networks. The overall framework of the proposed AutoML-based IDS is demonstrated in Fig. \ref{framework}, which comprises five stages: AutoDP, AutoFE, automated base model learning and selection, HPO, and automated model ensemble. During the initial stage, AutoDP, the input network traffic data undergoes pre-processing, where the proposed automated data balancing method identifies and addresses class-imbalance issues through the TVAE model to improve data quality. In the AutoFE stage, the most relevant features are automatically selected based on their importance scores calculated by the Gini index and entropy metrics using tree-based algorithms. This AutoFE process reduces data complexity and improves the generalization ability of the IDS model by minimizing noisy and redundant features. Subsequently, during the automated base model learning and selection stage, six tree-based ML algorithms (i.e., DT, RF, ET, XGBoost, LightGBM, and CatBoost) are trained and evaluated on the training set, and the top three best-performing models are automatically selected as the base models for further processing. In the HPO stage, the three selected ML models are further optimized through automated hyperparameter tuning or HPO using the BO-TPE method. In the automated model ensemble stage, the confidence values of all classes generated from the three optimized base models are integrated using the proposed OCSE model to obtain the final ensemble IDS model for final intrusion detection. 

Overall, this comprehensive AutoML-based framework enables the integration of advanced AutoML techniques across multiple stages,  collectively enhancing the detection capabilities and robustness of the proposed autonomous IDS model against various cyber threats for safeguarding next-generation networks.

\subsection{Automated Data Pre-Processing (AutoDP)}
Data pre-processing is an essential stage in the ML and data analytics pipeline, since it directly influences the quality of input data and, consequently, the performance of ML models \cite{auto4}. However, data pre-processing procedures can be tedious and time-consuming, often requiring massive human effort and expert knowledge. To address these challenges, Automated Data Pre-processing (AutoDP) has emerged as a critical component of AutoML that aims to automatically identify and address data quality issues in datasets, thereby ensuring that ML models can learn meaningful patterns from high-quality data \cite{myautoml}. 

In the proposed AutoML framework, the AutoDP component focuses on automated data balancing, a crucial aspect of data pre-processing that addresses class imbalance issues. Class imbalance is a common data quality issue in network data analytics and intrusion detection problems, as cyber-attacks or anomalies usually occur less frequently compared to benign or normal events, leading to a significant imbalance in class distribution. Class imbalance issues often bias ML models in IDS development, leading them to prioritize the detection of normal sample and common attacks and neglecting the less common but critical threats  \cite{myautoml}.

Data balancing techniques are designed to address class imbalance issues and can be classified into under-sampling and over-sampling techniques. Under-sampling methods alter the class distribution by eliminating instances from the majority classes to balance data, which can result in the loss of critical patterns of normal network activities \cite{balance}. On the other hand, over-sampling methods balance data by creating synthetic samples for the minority classes, which may slightly increase model training time but often outperform under-sampling methods. Therefore, over-sampling methods are considered in this research. 

Over-sampling methods can be broadly categorized into random and informed over-sampling methods \cite{myautoml}. Random over-sampling randomly replicates samples from the minority classes, while informed methods aim to generate higher-quality samples to improve the data balancing performance. In the proposed framework, the Tabular Variational Auto-Encoder (TVAE) \cite{tvae} method is used as an informed over-sampling method to synthesize minority class samples to balance tabular network data. TVAE is a DL model that extends the capabilities of the traditional autoencoder by incorporating probabilistic modeling and adapting to tabular data \cite{tvae}. TVAEs incorporate an encoder to project the input data $\bm{\mathit{x}}$ as a probability distribution in the latent space $\bm{\mathit{z}}$, denoted as $q(\bm{\mathit{z}}|\bm{\mathit{x}})$, and a decoder to reconstruct the data generation process as a probability distribution $p(\bm{\mathit{x}}|\bm{\mathit{z}})$. The objective function of TVAE is to maximize the Evidence Lower BOund (ELBO) on the log-likelihood of data, denoted by \cite{tvae2}:
\begin{equation}
ELBO = \mathbb{E}[\log p(\bm{\mathit{x}}|\bm{\mathit{z}})] - D_{KL}(q(\bm{\mathit{z}}|\bm{\mathit{x}}) || p(\bm{\mathit{z}}))
\end{equation}
where $D_{KL}$ is the Kullback-Leibler (KL) divergence, a measure of the difference between two probability distributions, and $\mathbb{E}$ denotes the expectation.

To automate the data balancing procedure, the proposed AutoDP method consists of two procedures: automated class-imbalance detection and automated data synthesis. In the automated class-imbalance detection procedure, the system calculates three metrics in the training set: the number of classes, the number of samples in each class, and the average number of samples per class. Based on these three metrics, the system will identify minority classes that have fewer than a certain threshold, defined as half of the average number of samples per class in the training set, in the proposed framework. If there are minority classes indicating class imbalance, the system will automatically perform automated data synthesis by synthesizing minority class samples using the TVAE method until the number of samples in each minority class reaches the threshold (half of the average number of samples). The details of the proposed TVAE-based automated data balancing method are provided in Algorithm \ref{algo:tvae}. Finally, a balanced training dataset is automatically obtained using the proposed AutoDP method, which involves generating high-quality minority class samples that better represent patterns of less common attacks, thereby assisting in effective IDS model development.

\begin{algorithm}[t!]
    {\small
    \caption{Automated Data Balancing Using Tabular Variational Auto-Encoder (TVAE)}
    \label{algo:tvae}
    \LinesNumbered
    \KwIn{ 
    $D_{train}$: the original training set of the dataset
    }
    \KwOut{
    $D_{train}^{bal}$: the balanced training set
    }
    $average\_samples$ = AverageNumberOfSamplesPerClass($D_{train}$)  \quad // Calculate the average number of samples per class \\
    $minority\_classes$ = IdentifyMinorityClasses($D_{train}$, $average\_samples$)  \quad // Identify classes with less than half the average samples \\
    $synthetic\_data$ = []  \quad // Initialize an empty list to store synthetic data \\
    \For{each class $cls$ in $minority\_classes$}{
        $cls\_samples$ = ExtractClassInstances($D_{train}$, $cls$)  \quad \quad \quad \quad \quad \quad // Extract instances of the minority class \\
        $num\_samples$ = $average\_samples$ - Count($cls\_samples$)  \quad \quad // Calculate the deficit in samples for the class \\
        $TVAE\_model$ = TrainTVAE($cls\_samples$)  \quad \quad \quad // Train the TVAE model on the minority class samples \\
        $new\_samples$ = GenerateSyntheticInstances($TVAE\_model$, $num\_samples$) \quad  // Generate synthetic instances to match the average class sample size \\
        $synthetic\_data$.append($new\_samples$)  \quad \quad \quad // Append the new synthetic instances to the synthetic data list \\
    }
    $D_{train}^{bal}$ = Concatenate($D_{train}$, $synthetic\_data$)   \quad  \quad // Concatenate the original and synthetic data to form a balanced dataset \\
    \KwRet $D_{train}^{bal}$
    }
\end{algorithm}

\subsection{Automated Feature Engineering (AutoFE)}
After the AutoDP stage, Automated Feature Engineering (AutoFE) is another crucial component of the proposed AutoML framework. Feature Engineering (FE) involves extracting and selecting the most informative and relevant features from a dataset, as the original features are often suboptimal for specific datasets \cite{auto2}. This process enhances the performance of ML models. AutoFE aims to automate the traditional FE process, minimizing human effort on FE tasks. In the proposed framework, AutoFE focuses on the Feature Selection (FS) process, aiming to identify and select the most relevant features to construct a highly efficient and accurate ML model. The proposed Automated FS (AutoFS) method is designed based on the feature importance scores generated by the tree-based algorithms used in the automated model learning process. 

To construct a DT in tree-based algorithms, features that result in significant reductions in Gini impurity or entropy will be assigned higher importance scores, as they have an important impact on the node-splitting process. Gini impurity and entropy are two common evaluation metrics to measure the impurity of nodes in DTs for classification problems to which intrusion detection problems belong \cite{gini}. The Gini index quantifies the impurity of a node by evaluating the probability of misclassifying a randomly selected element within that node. The Gini index is calculated as follows for a multi-class problem with $K$ classes \cite{gini}: 
\begin{equation}
\operatorname{Gini}\left(p_1, p_2, \ldots, p_K\right)=1-\sum_{i=1}^K p_i^2
\end{equation}

where $p_1, p_2, …, p_k$ indicate the proportions of classes $1, 2, …, k$.

Entropy is another impurity measure that calculates how much information is required to identify the class of a randomly selected element within a node \cite{gini}. The entropy for a multi-class problem with $K$ classes can be denoted by: 
\begin{equation}
\operatorname{Entropy}\left(p_1, p_2, \ldots, p_K\right)=-\sum_{i=1}^K p_i \log _2\left(p_i\right)
\end{equation}

In the proposed AutoFE framework, the FS process is automated by leveraging the power of tree-based models, as they can automatically calculate the feature importance scores during their training process. The specific procedures of this AutoFS process are as follows:
\begin{enumerate}
\item Train the six tree-based ML models (DT, RF, ET, XGBoost, LightGBM, and CatBoost) on the training set and evaluate their performance.
\item Obtain the feature importance scores generated from the top three best-performing models.
\item Calculate the average relative importance score for each feature across the top three best-performing models.
\item Rank the features from highest to lowest based on these average importance scores.
\item Select features from the top of this ranked list, accumulating their importance scores until the sum reaches a predefined threshold, $\alpha$ (default value is 0.9).
\item Generate the updated training and test sets using the newly generated feature set that comprises the selected features.

\end{enumerate}

By implementing the proposed AutoFE process, the most relevant features are selected based on their cumulative relative importance scores, ensuring a total of 90\%. Simultaneously, features with a cumulative importance score below 10\% are discarded, effectively reducing noise and computational complexity. The cumulative feature importance threshold, $\alpha$, can be tuned using the optimization method presented in Section \ref{s_hpo} to customize it for specific tasks or problems. The proposed AutoFE process helps to simplify the model, reduce the risk of overfitting, improve computational efficiency, and increase model interpretability.

\subsection{Automated Base Model Learning and Selection}
After the AutoDP and AutoFE procedures, the improved network traffic datasets are learners by supervised ML algorithms to train ML-based IDS that can detect various types of cyber-attacks. Six tree-based ML models, i.e., DT, RF, ET, XGBoost, LightGBM, and CatBoost, are built as base models to perform the initial intrusion detection. 

DT \cite{mth} is a fundamental ML algorithm that makes predictions by learning decision rules inferred from the input features. The decision rules are formed in a tree structure, where each internal node represents a test on a feature, each branch denotes a test outcome, and each leaf node contains a class label \cite{mth}. The DT algorithm recursively partitions data by selecting the best splitting rule, creating child nodes based on the chosen criterion, and repeating this process until a stopping condition is met.

RF \cite{rf} is an ensemble learning model constructed on multiple DT models. RF works by generating a set of DTs from randomly selected subsets of the training set and then aggregates the votes from the base DTs to decide the final result based on the majority voting rule. 

ET \cite{et} is another tree-based ensemble method constructed by combining multiple DTs. However, it randomizes both features and cut-point choices to construct completely randomized trees. As the splitting points in ETs are chosen randomly, the constructed trees in ETs are more diverse and less prone to over-fitting than in RF. 

XGBoost \cite{xgb} is an ensemble model built on the speed and performance of the Gradient-Boosted Decision Trees (GBDT) model. XGBoost distinguishes itself from traditional gradient boosting methods by incorporating a regularization term into the objective function, which effectively controls the model's complexity, smooths the final weights, and mitigates overfitting \cite{myautoml}. Additionally, XGBoost uses a second-order Taylor expansion to estimate the loss function, enabling an accurate model update and fast convergence. 

LightGBM \cite{lgb} is another improved version of the GBDT model with enhanced model performance and efficiency. Similar to other tree-based algorithms, LightGBM is constructed on an ensemble of DTs, but it introduces two advanced techniques, Gradient-based One-Side Sampling (GOSS) and Exclusive Feature Bundling (EFB) \cite{myautoml}. GOSS is a down-sampling approach that keeps instances with large gradients and randomly samples instances with small gradients to save model training time and memory. EFB groups mutually exclusive features into bundles as single features, reducing feature space dimensionality and improving training efficiency. By utilizing GOSS and EFB, LightGBM significantly reduces data size without losing critical information, preserving accuracy in the learning process and reducing computational cost.

CatBoost \cite{cb} is another GBDT-based algorithm that is particularly effective for datasets with categorical features. CatBoost distinguishes itself from traditional GBDT algorithms through three major innovations: symmetric trees, ordered boosting, and native feature support. Symmetric trees ensure that all the DTs in the model are symmetric, which simplifies the model and reduces the risk of overfitting. Ordered boosting is a novel boosting scheme that prevents overfitting on small-sized datasets. Native feature support allows CatBoost to handle categorical features for model performance enhancement natively.

The primary reasons for choosing these six tree-based algorithms as candidate base models are as follows:
\begin{enumerate}
\item RF, ET, XGBoost, LightGBM, and CatBoost are all ensemble models that combine multiple base DTs to improve model performance and robustness, and DT can serve as the baseline model for comparison. 
\item These methods are proficient at handling non-linear and high-dimensional data to which 5G network data belongs. 
\item They support parallel computation, which can significantly improve the training efficiency on large network datasets.
\item These tree-based ML algorithms offer the advantage of automatically calculating feature importance in their training process, which assists in efficient feature selection process in the proposed AutoFE method. 
\item These tree-based models incorporate randomness during their tree construction process, which enables the proposed framework to build a robust ensemble model with diverse base models and increased generalizability.

\end{enumerate}

After training and evaluating the performance of these six tree-based models on the training set, the three best-performing models based on their cross-validated F1-scores are automatically selected as the three base models to construct the ensemble model discussed in Section \ref{s_ensemble}.

\subsection{Hyper-Parameter Optimization (HPO)}
\label{s_hpo}
Tuning hyper-parameters is a crucial step in deploying an effective ML model to a particular problem or dataset. Hyper-parameters of a ML model determine its architecture and have a direct effect on the performance of this ML model. The process of using optimization methods to automatically tune and optimize these hyper-parameters is known as Hyper-Parameter Optimization (HPO). In HPO or AutoML tasks, Given the hyperparameter search space $X$,  the goal is to find the optimal hyper-parameter value or configuration $x^*$ that minimizes the objective function $f(x)$ \cite{tpe}:
\begin{equation}
x^*=\arg \min _{\boldsymbol{x} \in X} f(\boldsymbol{x})
\end{equation}

In the proposed AutoML framework, the important hyperparameters of the six tree-based algorithms are optimized during the HPO process. Utilizing terminology from the Scikit-Learn library, key hyperparameters for the Decision Tree (DT) include `\textit{max\_depth}', which sets the maximum tree depth; `\textit{min\_samples\_split}', specifying the minimum number of samples required to split a node; and `\textit{min\_samples\_leaf}', defining the minimum number of samples required at a leaf node. The `\textit{criterion}' hyperparameter allows selection between Gini impurity and entropy to measure splitting quality. As RF and ET are ensemble models built using DTs, they inherit these four critical hyperparameters from DT. Additionally, RF and ET include the `\textit{n\_estimators}' hyperparameter, which determines the number of trees in the ensemble and significantly influences model performance and efficiency.

The number of base trees and maximum tree depth are two crucial hyperparameters shared by XGBoost, LightGBM, and CatBoost. Additionally, since these three algorithms are gradient-boosting models, the learning rate is another critical hyperparameter that significantly impacts their learning speed and overall performance.

Among the various optimization methods for HPO tasks, Bayesian Optimization (BO) methods have proven to be efficient \cite{bo2}. BO leverages a posterior distribution, known as the surrogate, to describe the function under optimization. As more observations are made, the posterior distribution improves, which increases the certainty about promising regions in the parameter space worth exploring and the unpromising regions. Therefore, BO methods can determine future hyper-parameter evaluations based on the results of previous evaluations to avoid unnecessary model assessments \cite{mth}. 

The Tree Parzen Estimator (TPE) is a common surrogate for BO to model the evaluated configurations \cite{bo2}. BO with TPE, or BO-TPE, can handle a tree-structured hyper-parameter search space using Parzen estimators, also known as kernel density estimators (KDEs) \cite{tpe}. In BO-TPE, the hyper-parameter configuration space D is split into the better group $D^{(l)}$ and the worse group $D^{(g)}$ based on a top quantile $y^*$. The KDEs are estimated using a kernel function with a bandwidth that changes based on a provided dataset. The density functions of the configurations, modeled by TPE, can be denoted by \cite{tpe}:
\begin{equation}
p(\boldsymbol{x} \mid y, D)= \begin{cases}p\left(\boldsymbol{x} \mid y, D^{(l)}\right), & \text { if } y \leq y^*, \\ p\left(\boldsymbol{x} \mid y, D^{(g)}\right), & \text { if } y>y^*,\end{cases}
\end{equation}

The ratio between the two probability density functions is utilized to determine the new configurations for evaluation, facilitating the gradual identification of optimal configurations. BO-TPE is selected as the HPO method to tune and optimize the hyperparameters of the ML models in the proposed framework for the following reasons \cite{mth} \cite{myautoml}:
\begin{enumerate}
\item BO-TPE is effective for handling high-dimensional variables with multiple types, rendering it suitable for the tree-based ML methods utilized in the proposed framework, which involve numerous hyperparameters.
\item BO-TPE can handle tree-structured search spaces, enabling flexible and complex hyperparameter optimization, making it well-suited for the tree-based ML models employed in the proposed framework.
\item Unlike other HPO methods, like grid search, which treats each hyperparameter configuration independently and causes many unnecessary evaluations, BO-TPE enables more efficient HPO by exploring promising regions and determining new hyperparameter configurations based on previous evaluation results. 
\item BO-TPE has low time complexity of $O(n\log n)$, where $n$ is the number of hyperparameter configurations, which is much lower than other HPO methods, such as grid search, with time complexity of $O(n^k)$ \cite{myautoml}.

\end{enumerate}

By automatically tuning the hyperparameters of the three best-performing base ML models using BO-TPE, three optimized ML models with improved intrusion detection effectiveness are obtained for further analysis.

\subsection{Automated Model Ensemble}
\label{s_ensemble}

After selecting the top three best-performing tree-based models and optimizing their hyperparameters using BO-TPE, the three optimized models are utilized as base models to construct an ensemble model for further performance enhancement. Ensemble learning is an advanced technology that combines the prediction outcomes of multiple individual ML models to make final predictions \cite{a_ensemble}. Ensemble learning aims to improve model performance and generalizability by leveraging the collective knowledge of multiple models. 

In the last stage of the proposed AutoML framework, a novel Optimized Confidence-based Stacking Ensemble (OCSE) method is proposed to construct the final ensemble model by extending the traditional stacking ensemble strategy. Stacking is a widely-used ensemble learning method that comprises two layers of models. The first layer of stacking contains multiple trained base learners, and their output labels serve as the input for training a robust meta-learner in the second layer \cite{stacking}. 

Compared with the traditional stacking method, the proposed OCSE method introduces two additional strategies: confidence inputs and optimization. Firstly, the three optimized base ML models provide a probability distribution over the target classes for each sample, which indicates the confidence of the models’ prediction for this sample. The confidence values output by the base models are used as input to the meta-learner in the second layer of the proposed OCSE model.

Secondly, the best-performing base model from the six tree-based models, based on the cross-validated F1-scores, is selected to construct the meta-learner. Its hyperparameters are optimized by BO-TPE to obtain the optimized meta-learner, following the same HPO process used for the base models, as described in Section \ref{s_hpo}. The specifications of the proposed OCSE method and the entire AutoML framework are illustrated in Algorithm  \ref{algo:automl}. 

The computational complexity of the proposed OCSE model is $O(ncmh)$, where $n$ is the number of samples, $c$ is the number of unique classes, $m$ is the number of base models, and $h$ is the number of hyperparameter configurations of the meta-learner. In the proposed framework, $c$, $m$, and $h$ are all relatively small numbers.

Compared with other ensemble techniques, the proposed OCSE method presents the following advantages: 
\begin{enumerate}
\item \textit{Utilization of Confidence}: Unlike many existing ensemble techniques, such as bagging, boosting, and traditional stacking, which solely rely on the predicted labels to construct the ensemble model, the proposed OCSE method utilizes the confidence of all classes as input features, which provides more comprehensive information about the certainty of base model's predictions, resulting in more informed and robust ensemble predictions.
\item \textit{Automated and Optimized Models}: The proposed OCSE method automatically selects the best-performing base model as the second-layer meta-learner and tunes its hyperparameters, resulting in an optimized final learner capable of achieving the optimal overall performance. The automation process also reduces the need for manual effort and saves time in model development.
\item \textit{Flexibility}: OCSE is a flexible method in which both the base models and the meta-learner can be replaced with other ML algorithms to adapt to a wide range of tasks.
\end{enumerate}

\begin{algorithm}[t!]
    {\small
    \caption{The Proposed AutoML-based IDS Framework Involving The OCSE Algorithm}
    \label{algo:automl}
    \LinesNumbered
    \KwIn{ 
    $D_{train}$, $D_{test}$
    }
    \KwOut{
    $L_{pred}$: Predicted labels for normal and attack samples
    }
    \textbf{Stage 1: Automated Data Preprocessing (AutoDP)}\\
    \quad $D_{train}^{bal}$ = BalanceData($D_{train}$) using Algorithm 1 \qquad\qquad\qquad\qquad// Address class imbalance using Algorithm 1 and TVAE \\
    \textbf{Stage 2: Automated Feature Engineering and Base Model Selection (AutoFE)}\\
    \quad Train and evaluate $BM_i$ (DT, RF, ET, XGBoost, LightGBM, CatBoost) on $D_{train}^{bal}$  \qquad\qquad\qquad \qquad// Train and evaluate base models using cross-validation\\
    \quad $m_i$ = Metrics($BM_i$) for each model $i \in \{1,2,3,4,5,6\}$ \qquad\qquad\qquad// Compute performance metrics for model evaluation\\
    \quad $M_1$, $M_2$, $M_3$ = SelectTopModels($m_i$, 3) \qquad\qquad\qquad// Select the three best-performing models based on metrics\\
    \quad $F_1$, $F_2$, $F_3$ = FeatureImportances($M_1$, $M_2$, $M_3$) \qquad\qquad\qquad\qquad// Calculate feature importance for selected models\\
    \quad $F_{avg} = \frac{F_1 + F_2 + F_3}{3}$ \qquad\qquad// Average the feature importance scores to improve generalization\\
    \quad $F_s$ = SelectFeatures($F_{avg}$, $\alpha = 90\%$) \qquad\qquad// Select most important features that cumulatively meet importance threshold $\alpha$\\
    \quad $M_1'$, $M_2'$, $M_3'$ = RetrainModels($M_1$, $M_2$, $M_3$, $F_s$) \qquad\qquad// Retrain models using selected features\\
    \textbf{Stage 3: Hyperparameter Optimization (HPO) using BO-TPE}\\
    \quad $M_1 = (H_1, S_1)$, $M_2 = (H_2, S_2)$, $M_3 = (H_3, S_3)$ \qquad// Configure search spaces for the hyperparameters of three selected base models\\
    \quad $h_1^*$, $h_2^*$, $h_3^*$ = OptimizeHP($S_1$, $S_2$, $S_3$) using BO-TPE \qquad\qquad\qquad// Optimize hyperparameters using Bayesian Optimization\\
    \quad $M_1''$, $M_2''$, $M_3''$ = GenerateOptimizedModels($h_1^*$, $h_2^*$, $h_3^*$) \qquad\qquad// Generate models with optimized hyperparameters\\
    \textbf{Stage 4: Model Ensemble using Optimized Confidence-based Stacking Ensemble (OCSE)}\\
    \quad $P_1$, $P_2$, $P_3$ = ConfidenceValues($M_1''$, $M_2''$, $M_3''$, $D_{train}^{bal}$, $F_s$) \qquad\qquad// Retrieve confidence values from optimized models\\
    \quad $M$ = TrainMetaLearner($P_1$, $P_2$, $P_3$, $M_1$) \qquad\qquad// Use the best-performing ML model to train a meta-learner on model confidence values\\
    \quad $M'$ = OptimizeMetaLearner($M$, BO-TPE) \qquad\qquad// Optimize the meta-learner using BO-TPE\\
    \quad $P_1'$, $P_2'$, $P_3'$ = TestConfidences($M_1''$, $M_2''$, $M_3''$, $D_{test}$)\qquad // Retrieve confidence values for test data\\
    \quad $L_{pred}$ = FinalPredictions($M'$, $P_1'$, $P_2'$, $P_3'$) \qquad\qquad// Predict final labels using the optimized meta-learner based on confidence values\\
    \KwRet $L_{pred}$
    }
\end{algorithm}

Overall, with the use of the novel OCSE method and all the other critical components described in this section, the proposed AutoML-based IDS framework can automatically generate an optimized ensemble model for effective and robust intrusion detection, serving as a key component for autonomous cybersecurity solutions.

\section{Performance Evaluation}
\subsection{Experimental Setup}
The proposed AutoML-based IDS framework was developed in Python by extending the Scikit-Learn \cite{sklearn}, Xgboost \cite{xgb}, Lightgbm \cite{lgb}, Catboost \cite{cb}, Synthetic Data Vault (SDV) \cite{sdv}, and Hyperopt \cite{hyperopt} libraries. The experiments were performed on a Dell Precision 3630 computer equipped with an i7-8700 processor and 16 GB of memory, which served as the server machine in 5G and potential 6G networks.

To evaluate the proposed AutoML-based IDS framework, two public benchmark network traffic datasets, namely CICIDS2017 \cite{cic} and 5G-NIDD \cite{5gnidd}, are utilized in the experiments. The CICIDS2017 dataset is one of the most comprehensive public cybersecurity datasets, created by simulating a real-world network environment and involves six primary types of attacks: DoS, botnets, brute force, infiltration, port scan, and web attacks \cite{mth}. The diverse attack scenarios and comprehensive feature set of the CICIDS2017 dataset make it suitable for network security applications. The 5G-NIDD dataset is one of the most state-of-the-art network security datasets developed in December 2022 \cite{5gnidd}. This dataset was generated by capturing network traffic within a 5G testbed under diverse DoS and port scan cyber-attacks. The 5G-NIDD dataset is particularly suitable for our research, as it specifically targets 5G networks and enables the detection of new and sophisticated cyber-attacks.

To develop and evaluate the proposed AutoML-based IDS model, both cross-validation and hold-out validation methods are used in the experiments. The model training and optimization process utilizes five-fold cross-validation to automatically generate the optimized ensemble model, while the final model produced by the proposed AutoML framework is evaluated using an unseen test set, which is split from an 80\%/20\% hold-out validation during the initial stage of data pre-processing.

Due to the inherent class imbalance issues in network intrusion detection datasets, four model performance metrics—accuracy, precision, recall, and F1-scores—are considered collectively in the experiments. The F1-score is utilized as the primary performance metric in the performance-based automated model selection and tuning process of the proposed AutoML framework, as it offers a balanced view of anomaly detection results by calculating the harmonic mean of recall and precision. Additionally, the model execution time, involving the training and inference time of the final OCSE model, is utilized to assess the model's efficiency.

\subsection{Experimental Results and Discussion}
\begin{figure}
     \centering
     \includegraphics[width=\columnwidth]{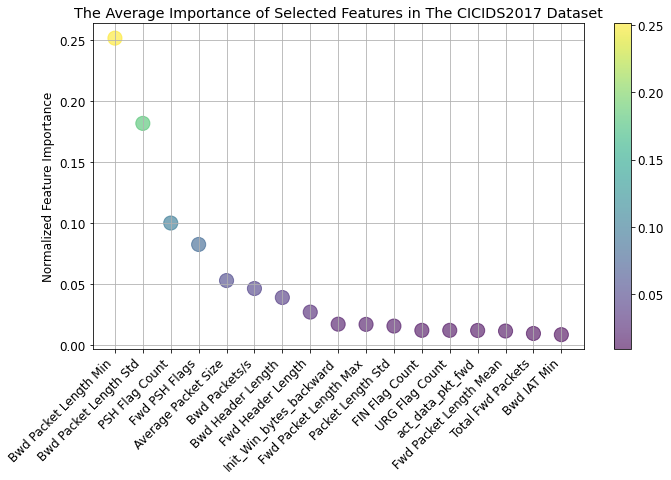}
     \caption{The average importance scores of the selected features in the CICIDS2017 dataset (cumulative relative importance reaching 90\%).} 
     \label{fs_cic}
\end{figure}

\begin{figure}
     \centering
     \includegraphics[width=\columnwidth]{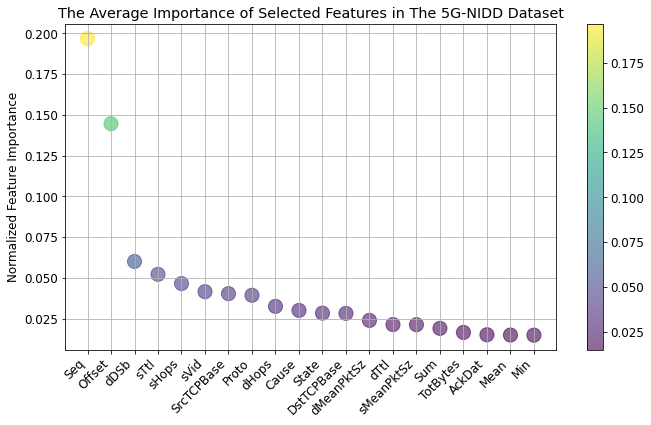}
     \caption{The average importance scores of the selected features in the 5G-NIDD dataset (cumulative relative importance reaching 90\%).} 
     \label{fs_5g}
\end{figure}

As described in Section \ref{S3}, the proposed AutoML-based IDS comprises five critical stages: AutoDP, AutoFE, automated base model learning and selection, HPO, and automated model ensemble. Initially, the proposed TVAE-based automated data balancing method applied in the AutoDP stage automatically balances the distributions of two datasets, CICIDS2017 and 5G-NIDD, to prevent model bias. During the AutoFE stage, important features are selected based on the average importance scores obtained from the three best-performing ML models, where the accumulative importance score reaches the threshold $\alpha=90\%$. These selected features, along with their relative importance scores for the CICIDS2017 and the 5G-NIDD datasets, are illustrated in Figs. \ref{fs_cic} and \ref{fs_5g}, respectively. Subsequently, the three top-performing ML models—RF, XGBoost, and LightGBM—are optimized by tuning their hyperparameters using BO-TPE. The hyperparameters tuned, their search spaces, and the optimal values obtained for these hyperparameters for both datasets are detailed in Table \ref{t_hpo}. Finally, the three optimized base ML models are integrated using the proposed OCSE model to automate the model ensemble, improving the decision-making effectiveness in intrusion detection.
 
The performance of the proposed AutoML-OCSE model and several state-of-the-art methods in the literature is provided in Table \ref{ids_table1} for the CICIDS2017 dataset and Table \ref{ids_table2} for the 5G-NIDD dataset. The performance is evaluated based on the metrics: accuracy, precision, recall, F1-score, training time, and average test time per sample. 
Firstly, as shown in Table \ref{ids_table1}, results on the CICIDS2017 dataset indicate that RF, XGBoost, and LightGBM perform better than the other three base models, DT, ET, and CatBoost. Hence, these three ML models are selected as the base models of the proposed AutoML-OCSE framework. 
After optimizing the hyperparameters of the three selected base models (as detailed in Table \ref{t_hpo}) and integrating their outputs using the proposed OCSE ensemble method, the final AutoML-OCSE model outperforms all the compared methods in the literature \cite{mth} \cite{cic} \cite{ml1} \cite{ids3} - \cite{ids6}. The proposed method achieves the highest metrics on the CICIDS2017 dataset, with accuracy, precision, recall, and F1-score of 99.806\%, 99.806\%, 99.806\%, and 99.804\%, respectively.

Furthermore, the average test time per sample of the proposed AutoML-OCSE method is the fastest among the compared methods \cite{mth} \cite{cic} \cite{ml1} \cite{ids3} - \cite{ids6}, highlighting its efficiency on network traffic datasets. Compared with state-of-the-art methods on the CICIDS2017 dataset, AutoML-OCSE demonstrates notable improvements in both accuracy and inference efficiency. These enhancements are attributed primarily to the AutoDP, AutoFE, HPO, and automated model ensemble procedures, which collectively enhance data quality, optimize machine learning models, and reduce feature and model complexity.

Similarly, as indicated in Table \ref{ids_table2}, the proposed AutoML-OCSE method outperforms all other compared methods \cite{mth} \cite{rf} - \cite{cb} \cite{cic} \cite{ids2} on the 5G-NIDD dataset, achieving the highest accuracy, precision, recall, and F1 score, all at 99.956\%. In terms of average test time per sample, the AutoML-OCSE method matches the fastest time set by the DT method, demonstrating an exceptional balance between performance and efficiency.

\begin{table} [!t]
\caption{The HPO configuration of the three best-performing models.}
\setlength\extrarowheight{1pt}
\centering
\footnotesize
\begin{tabular}{|>{\centering\arraybackslash}p{0.93cm}|>{\centering\arraybackslash}p{1.75cm}|>{\centering\arraybackslash}p{1.45cm}|>{\centering\arraybackslash}p{1.35cm}|>{\centering\arraybackslash}p{1.35cm}|}
\hline
\textbf{Model}            & \textbf{Hyperparameter Name} & \textbf{Configuration Space} & \textbf{Optimal Value on CICIDS2017} & \textbf{Optimal Value on 5G-NIDD}  \\ 
\hline
\multirow{5}{*}{RF}       & n\_estimators                & {[}50,500]                   & 420                                  & 370                                \\ 
\cline{2-5}
                          & max\_depth                   & {[}5,50]                     & 36                                   & 42                                 \\ 
\cline{2-5}
                          & min\_samples\_split          & {[}2,11]                     & 7                                    & 4                                  \\ 
\cline{2-5}
                          & min\_samples\_leaf           & {[}1,11]                     & 2                                   & 5                                  \\ 
\cline{2-5}
                          & criterion                    & {[}’gini’, ’entropy’]        & ’entropy’                            & ’gini’                          \\ 
\hline
\multirow{5}{*}{XGBoost}  & n\_estimators                & {[}50,500]                 & 450                                  & 310                                \\ 
\cline{2-5}
                          & max\_depth                   & {[}5,50]                     & 36                                    & 27                                  \\ 
\cline{2-5}
                          & learning\_rate               & (0, 1)                  & 0.78                                  & 0.67                               \\ 
\cline{2-5}
                          & gamma                        & (0, 5)                       & 0.4                                  & 0.3                                \\ 
\cline{2-5}
                          & subsample                    & (0.5, 1)                     & 0.7                                  & 0.75                               \\ 
\hline
\multirow{5}{*}{LightGBM} & n\_estimators                & {[}50,500]                   & 380                                  & 340                                \\ 
\cline{2-5}
                          & max\_depth                   & {[}5,50]                     & 42                                   & 34                                 \\ 
\cline{2-5}
                          & learning\_rate               & (0, 1)                       & 0.918                                & 0.784                              \\ 
\cline{2-5}
                          & num\_leaves                  & {[}100,2000]                 & 900                                 & 1100                               \\ 
\cline{2-5}
                          & min\_child\_samples          & {[}10,50]                    & 42                                   & 28                                 \\ 
\hline
\end{tabular}
\label{t_hpo}%
\end{table}

\begin{table}[!t]
\caption{Model Performance Comparison on CICIDS2017.}
\centering
\setlength\extrarowheight{1pt}
\scalebox{0.75}{
\begin{tabular}{|>{\centering\arraybackslash}p{6em}|>{\centering\arraybackslash}p{3.7em}|>{\centering\arraybackslash}p{3.7em}|>{\centering\arraybackslash}p{2.8em}|>{\centering\arraybackslash}p{2.8em}|>{\centering\arraybackslash}p{3.6em}|>{\centering\arraybackslash}p{5.5em}|}
\hline
\textbf{Method} & \textbf{Accuracy (\%)} & \textbf{Precision (\%)} & \textbf{Recall (\%)} & \textbf{F1 (\%)} & \textbf{Training Time (s)} & \textbf{Avg Test Time Per Sample (ms)} \\
\hline
KNN \cite{cic} & 96.3 & 96.2 & 93.7 & 96.3 & 0.007 & 0.678 \\
\hline
DT \cite{mth} & 99.612 & 99.612 & 99.612 & 99.608 & 0.6 & 0.0008 \\
\hline
RF \cite{rf} & 99.718 & 99.718 & 99.718 & 99.714 & 38.5 & 0.014 \\
\hline
ET \cite{et} & 99.245 & 99.252 & 99.245 & 99.243 & 3.5 & 0.012 \\
\hline
XGBoost \cite{xgb} & 99.757 & 99.757 & 99.757 & 99.755 & 11.0 & 0.001 \\
\hline
LightGBM \cite{lgb} & 99.770 & 99.770 & 99.770 & 99.769 & 2.0 & 0.004 \\
\hline
CatBoost \cite{cb} & 99.559 & 99.559 & 99.559 & 99.553 & 4.6 & 0.008 \\ 
\hline
KNN-AIDS \cite{ml1} & 99.52 & 99.49 & 99.52 & 99.49 & - & - \\
\hline
DL-LSTM \cite{ids3} & 99.32 & 99.32 & 99.32 & 99.32 & - & - \\
\hline
PyDSC-IDS \cite{ids4} & 97.60 & 90.73 & 97.81 & 94.13 & - & - \\
\hline
OE-IDS \cite{ids5} & 98.0 & 97.3 & 96.0 & 96.7 & - & - \\
\hline
PSO-DL \cite{ids6} & 98.95 & 95.82 & 95.81 & 95.81 & - & - \\
\hline
\textbf{Proposed AutoML-OCSE} & \textbf{99.806} & \textbf{99.806} & \textbf{99.806} & \textbf{99.804} & 35.6 & \textbf{0.0007} \\
\hline
\end{tabular}
}
\label{ids_table1}%
\end{table}

\begin{table}[!t]
\caption{Model Performance Comparison on 5G-NIDD.}
\centering
\setlength\extrarowheight{1pt}
\scalebox{0.75}{
\begin{tabular}{|>{\centering\arraybackslash}p{6em}|>{\centering\arraybackslash}p{3.7em}|>{\centering\arraybackslash}p{3.7em}|>{\centering\arraybackslash}p{2.8em}|>{\centering\arraybackslash}p{2.8em}|>{\centering\arraybackslash}p{3.6em}|>{\centering\arraybackslash}p{5.5em}|}
\hline
\textbf{Method} & \textbf{Accuracy (\%)} & \textbf{Precision (\%)} & \textbf{Recall (\%)} & \textbf{F1 (\%)} & \textbf{Training Time (s)} & \textbf{Avg Test Time Per Sample (ms)} \\
\hline
KNN \cite{cic} & 99.007 & 99.008 & 99.007 & 99.007 & 0.006 & 0.704 \\
\hline
DT \cite{mth} & 99.926 & 99.926 & 99.926 & 99.926 & 0.25 & 0.0006 \\
\hline
RF \cite{rf} & 99.942 & 99.942 & 99.942 & 99.942 & 3.2 & 0.009 \\
\hline
ET \cite{et} & 99.926 & 99.926 & 99.926 & 99.926 & 2.3 & 0.010 \\
\hline
XGBoost \cite{xgb} & 99.942 & 99.942 & 99.942 & 99.942 & 7.1 & 0.0009 \\
\hline
LightGBM \cite{lgb} & 99.942 & 99.942 & 99.942 & 99.942 & 1.9 & 0.008 \\
\hline
CatBoost \cite{cb} & 99.918 & 99.918 & 99.918 & 99.918 & 29.7 & 0.009 \\
\hline
Embeddings \& FC \cite{ids2} & 99.123 & 99.019 & 98.316 & 98.666 & - & - \\
\hline
\textbf{Proposed AutoML-OCSE} & \textbf{99.956} & \textbf{99.956} & \textbf{99.956} & \textbf{99.956} & 24.3 & \textbf{0.0006} \\
\hline
\end{tabular}
}
\label{ids_table2}%
\end{table}

Regarding the intrusion detection performance of the proposed AutoML-OCSE on the CICIDS2017 and 5G-NIDD datasets, while its accuracy and F1-score are only slightly higher than those of the best-performing base ML models, this is primarily attributed to the simplicity of these datasets, where many base ML models can achieve over 99\% accuracy and F1-score. In real-world scenarios, where the complexity and variability of network traffic datasets are usually higher than those of public benchmarks, the proposed AutoML-based IDS is expected to demonstrate more significant improvements. This enhancement is due to every component of the proposed framework, including AutoDP, AutoFE, automated model selection, HPO, and automated model ensemble, each contributing to the overall enhancement of intrusion detection performance. Additionally, considering the error rate reduction, the proposed AutoML-OCSE method achieves significant decreases of approximately $15.65\%$ and $24.13\%$ in the error rates on the CICIDS2017 and 5G-NIDD datasets, respectively, calculated as $\frac{99.806\% - 99.770\%}{100\% - 99.770\%}$ and $\frac{99.956\% - 99.942\%}{100\% - 99.942\%}$. 
Furthermore, without the proposed automated model selection method, researchers might resort to selecting ML models either randomly or based on personal experience, which may not always lead to choosing the best-performing base ML model. This limitation further highlights the significant potential for improvement offered by the proposed AutoML method, which can automatically select, optimize, and integrate the best-performing ML models. Consequently, the proposed AutoML-based IDS can achieve substantial improvements over traditional cybersecurity methods through its autonomous cybersecurity strategies.

On the other hand, although the AutoML-OCSE model takes longer to train than certain base models, such as DT, ET, and LightGBM, its training time is still shorter than that of some other models, like CatBoost. This efficiency is due in part to its AutoFE process, which reduces data dimensionality and model complexity. Moreover, the improvement in accuracy, precision, recall, and F1-score justifies the slightly increased training time. Furthermore, the AutoML-OCSE model achieves the fastest average inference time per sample by constructing a stacking ensemble model based on confidence values rather than the original high-dimensional dataset, making it highly suitable for real-time network data analytics and intrusion detection applications. In network applications, low inference time is often more crucial than low training time, as model training typically occurs on cloud servers with ample computational resources, while model predictions are performed on edge or local devices with limited computational capabilities in many scenarios.

Overall, the performance results demonstrate the effectiveness and efficiency of the proposed AutoML-OCSE method. It integrates the strengths of various base ML models, automates tedious ML tasks through AutoML, and achieves high detection performance via an optimized ensemble strategy. Therefore, the proposed AutoML-OCSE method and the AutoML framework can serve as powerful autonomous cybersecurity solutions for intrusion detection in 5G and potential 6G networks.

\section{Conclusion}
The advent of 5G and the impending transition to 6G networks have underscored the importance of ZTNs in achieving network automation. However, the increased connectivity and complexity of these networks have also escalated cybersecurity risks, making the development of effective and autonomous cybersecurity mechanisms a critical necessity. In this paper, we propose an AutoML-based Intrusion Detection System (IDS) to achieve autonomous cybersecurity for future networks. By introducing AutoDP, AutoFE, automated base model learning and selection, HPO, and automated model ensemble components, the proposed AutoML-based IDS can automatically generate an optimized ensemble model for accurate intrusion detection. This paper also proposed a novel TVAE-based automated data balancing method and a novel OCSE model to improve the AutoML procedures. Through the experiments, the proposed AutoML-based IDS achieves high F1-scores of 99.804\% and 99.956\% on two public benchmark network security datasets: the CICIDS2017 and 5G-NIDD datasets. This illustrates the effectiveness of the proposed autonomous IDS framework in achieving autonomous cybersecurity. In future work, the IDS framework will be extended to involve automated model updating using continual learning and drift adaptive methods in dynamic networking environments.

\bibliographystyle{ACM-Reference-Format}
\balance
\bibliography{main}

\end{document}